\documentclass[a4paper]{article}

\usepackage{graphicx,verbatim,color,latexsym,amsmath,amssymb,multirow}
\usepackage{url}

\usepackage{algorithm}
\usepackage{clrscode}
\usepackage{xspace}

\fontfamily{ptm}

\newenvironment{proof}{\noindent{\em Proof.~}}{\hfill$\Box$\medskip}
\newenvironment{example}{\noindent{\em Example.~}}{\hfill$\Box$\medskip}
\newenvironment{theorem}{\noindent{\em Theorem.~}}{\hfill$\Box$\medskip}

\begin{document}

\def\baselinestretch{0.98}

\title{Algorithms for Weighted Boolean Optimization}

\author{Vasco Manquinho$^1$, Joao Marques-Silva$^2$, Jordi Planes$^3$\\[1em]
\small$^1$ IST/UTL - INESC-ID, \texttt{vasco.manquinho@inesc-id.pt} \\
\small$^2$ University College Dublin, \texttt{jpms@ucd.ie} \\
\small$^3$ Universitat de Lleida, \texttt{jplanes@diei.udl.cat}
}

\maketitle

\begin{abstract}
The Pseudo-Boolean Optimization (PBO) and Maximum Satisfiability
(MaxSAT) problems are natural optimization extensions of Boolean
Satisfiability (SAT).
In the recent past, different algorithms have been proposed for PBO
and for MaxSAT, despite the existence of straightforward mappings from
PBO to MaxSAT and vice-versa. This papers proposes Weighted Boolean
Optimization (WBO), a new unified framework that aggregates and
extends PBO and MaxSAT. In addition, the paper proposes a new
unsatisfiability-based algorithm for WBO, based on recent
unsatisfiability-based algorithms for MaxSAT. Besides standard MaxSAT, 
the new algorithm can also be used to solve weighted MaxSAT and PBO,
handling pseudo-Boolean constraints either natively or by translation 
to clausal form. Experimental results illustrate that unsatisfiability-based
algorithms for MaxSAT can be orders of magnitude more efficient than
existing dedicated algorithms. Finally, the paper illustrates how
other algorithms for either PBO or MaxSAT can be extended to WBO.
\end{abstract}

\section{Introduction} \label{sec:intro}

In the area of Boolean-based decision and optimization procedures,
natural extensions of Boolean Satisfiability~(SAT) include Maximum
Satisfiability~(MaxSAT)~\cite{borchers-jco99} and Pseu\-do-Boolean
Optimization~(PBO)~\cite{barth-95}.
Algorithms for MaxSAT and PBO have been the subject of significant
improvements over the last few years. This in turn, motivated the use
of both PBO and, more recently, of MaxSAT in a number of practical
applications.
Interestingly, albeit there are simple translations from any MaxSAT
variant to PBO and vice-versa (by encoding to
CNF)~\cite{kas-iccad02,HLO08}, algorithms for MaxSAT and PBO have
evolved separately, and often use fairly different algorithmic
organizations.
Nevertheless, there exists work that acknowledges this relationship
and algorithms that can solve instances of MaxSAT and of
PBO have already been proposed~\cite{kas-iccad02,HLO08}.

Recent work has provided more alternatives for solving either MaxSAT
or PBO, by using SAT solvers and the identification of unsatisfiable
sub-formulas~\cite{FM06,jpms-date08}. However, the proposed algorithms
were restricted to the plain and partial variants of MaxSAT and to a
restricted form of Binate Covering for PBO.
This paper extends this recent work in a number of directions. First,
the paper proposes a simple algorithm for (Partial) Weighted MaxSAT,
using unsatisfiable sub-formula identification. Second, the paper
generalizes MaxSAT and PBO by introducing Weighted Boolean
Optimization (WBO), a new modeling framework for solving linear
optimization problems over Boolean domains. Third, the paper shows
how to extend the unsatisfiability-based algorithm for MaxSAT for
solving WBO problems. Finally, the paper suggests how other algorithms
can be used for solving WBO.
Besides the proposed contributions, the paper also provides empirical
evidence that unsatisfiability-based MaxSAT and WBO solvers can
outperform state-of-the-art solvers on problem instances from practical
problems.

The paper is organized as follows. Section~\ref{sec:prelim} provides a
brief overview of the topics addressed in the paper, namely MaxSAT,
PBO, translations from MaxSAT to PBO and vice-versa, and
unsatisfiability-based algorithms for MaxSAT.
Section~\ref{sec:wmxsat} details an algorithm for (Partial) Weighted
MaxSAT based on unsatisfiable sub-formula identification. Next,
Section~\ref{sec:wbo} introduces Weighted Boolean Optimization (WBO), 
and shows how to extend the algorithm of Section~\ref{sec:wmxsat} to WBO.
Section~\ref{sec:res} analyzes the experimental results, obtained on
representative classes of problem instances. Section~\ref{sec:relw}
overviews related work, and Section~\ref{sec:conc} concludes the paper.

\section{Preliminaries}
\label{sec:prelim}

This section briefly introduces the Maximum Satisfiability (MaxSAT)
problem and its variants, as well as the Pseudo-Boolean Optimization
(PBO) problem. The main approaches used by state-of-the-art solvers
are summarized. Moreover, translation procedures from MaxSAT to PBO
and vice-versa are overviewed. Finally, unsatisfiability-based MaxSAT
algorithms are surveyed, all of which the paper uses in later
sections.

\subsection{Maximum Satisfiability}
\label{sec:maxsat}

Given a CNF formula $\varphi$, the Maximum Satisfiability (MaxSAT)
problem can be defined as finding an assignment that maximizes the 
number of satisfied clauses (which implies that the assignment 
minimizes the number of unsatisfied clauses).
Besides the classical MaxSAT problem, there are also three well-known
variants of MaxSAT: weighted MaxSAT, partial MaxSAT and weighted 
partial MaxSAT.
All these formulations have been used in a wide range of practical 
applications, namely scheduling, FPGA routing~\cite{kas-tcad03}, 
design automation~\cite{kas-fmcad07}, among others.

A partial CNF formula is described as the conjunction of
two CNF formulas $\varphi_h$ and $\varphi_s$, where $\varphi_h$ 
represents the \emph{hard} clauses and $\varphi_s$ represents the
\emph{soft} clauses. The \emph{partial} MaxSAT problem consists
in finding an assignment to the problem variables such that all
hard clauses ($\varphi_h$) are satisfied, and the number of satisfied 
soft clauses ($\varphi_s$) is maximixed.

A weighted CNF formula is a set of weighted clauses.
A weighted clause is a pair $(\omega,c)$, where $\omega$ is a classical 
clause and $c$ is a natural number corresponding to the cost of 
unsatisfying $\omega$. Given a weighted CNF formula, the \emph{weighted} 
MaxSAT problem consists in finding an assignment to the problem
variables such that the total weight of the satified clauses is
maximized (which implies that the total weight of the unsatisfied
clauses is minimized).

A weighted partial CNF formula is the conjunction of a weighted CNF
formula (soft clauses) and a classical CNF formula (hard clauses).
The \emph{weighted partial} MaxSAT problem consists in
finding an assignment to the variables such that all hard clauses are
satisfied and the total weight of satisfied soft clauses is 
maximized. Observe that, for both partial MaxSAT and 
weighted partial MaxSAT, hard clauses can also be represented
as weighted clauses: one can consider that the weight is greater than 
the sum of the weights of the soft clauses.

Starting with the seminal work of Borchers and Furman~\cite{borchers-jco99}, 
there has been an increasing interest in developing efficient MaxSAT
solvers. Following such work, two branch and bound based solvers 
have been developed: (i)~MaxSatz~\cite{LMP07}, the first solver to implement 
a unit propagation based lower bound and a failed literal based lower bound,
both closely linked with a set of inference rules;
(This solver has been extended into several solvers: 
IncMaxSatz~\cite{lin-ijcai07},
WMaxSatz~\cite{argelich-ncp07}, 
WMaxSatz\_icss~\cite{darras-cp07}.) 
(ii)~MiniMaxSAT~\cite{HLO08}, a solver created on top of MiniSAT with 
MaxSAT resolution~\cite{BLM07} applied over an unsatisfiable sub-formula 
detected by the unit propagation based lower bound.
A different approach has been the conversion of MaxSAT into 
a different formalism. The most notable works using this approach have 
been: Toolbar~\cite{heras-aij08}, a weighted CSP solver which converts 
MaxSAT instances into a weighted constraint network;
SAT4J MAXSAT~\cite{sat4j}, a solver which iteratively converts a MaxSAT 
instance into a PBO instance;
Clone~\cite{pipatsrisawat08solving} and sr(w)~\cite{RG07}, solvers which
convert a MaxSAT instance into a deterministic decomposable negation 
normal form (d-DNNF) instance; and
MSUnCore~\cite{jpms-date08}, a solver which solves MaxSAT using the 
unsatisfiable cores detected by iteratively encoding the problem 
instance into SAT.
In the Max-SAT Evaluations~\cite{maxsat-evaluation}, this latter
approach has been shown to be effective for industrial problems.

\subsection{Pseudo-Boolean Optimization}
\label{sec:pbo}

The Pseudo-Boolean Optimization (PBO) problem is another extension of
SAT where constraints can be any linear inequality with integer
coefficients (also known as pseudo-Boolean constraints) defined over 
the set of problem variables. The objective in PBO is to find an
assignment to problem variables such that all problem constraints
are satisfied and the value of a linear objective function is
optimized. Any pseudo-Boolean formulation can be easily translated
into a normal form~\cite{barth-95} such that all integer coefficients
are non-negative. 
\begin{equation}
  \begin{array}{lll}
    \mbox{minimize } & \sum\limits_{j \in N} c_j \cdot x_j \\
    \mbox{subject to } & \sum\limits_{j \in N} a_{ij} l_j \ge b_i, \\
    \mbox{ } & l_j \in \{x_j, \bar{x}_j\}, x_j \in \{0,1\},
    a_{ij}, b_i, c_j \in \mathbf{N}_0^+ \\
  \end{array}
  \label{eq:pbo_definition}
\end{equation}

Almost all algorithms to solve PBO rely on the generalization of the 
most effective techniques already used in SAT solvers, namely Boolean 
Constraint Propagation, conflict-based learning and conflict-directed 
backtracking~\cite{jpms-tcad02,chai-dac03}.
Nevertheless, there are several approaches to solve PBO formulations. 
The most common using SAT solvers is to make a linear search on the 
value of the objective function. The idea is to generalize SAT algorithms
to deal natively with pseudo-Boolean constraints~\cite{barth-95} and 
whenever a solution for the problem constraints is found, a new constraint 
is added such that only solutions with a lower value for the objective 
function can be accepted. The algorithm finishes when the solver cannot 
improve on the last solution found, therefore proving its optimality.

Another common approach is branch and bound, where lower bounding
procedures to estimate the value of the objective function are 
used. Several lower bounding procedures have been proposed,
namely Maximum Independent Set of constraints~\cite{coudert-96}, 
Linear Programming Relaxation~\cite{liao-97,jpms-date05}, among
others~\cite{jpms-date05}. 
There are also algorithms that encode pseudo-Boolean constraints into
propositional clauses~\cite{warners-ipl98,roussel-jsat06,een-jsat06} 
and solve the problem by subsequently using a SAT solver. This 
approach has been proved to be very effective for several problem 
sets, in particular when the clause encoding is not much larger 
than the original pseudo-Boolean formulation.

\subsection{Translations between MaxSAT and PBO}
\label{sec:ms-pbo}

Although MaxSAT and PBO are different formalisms, it is possible to 
encode any MaxSAT instance into a PBO instance and
vice-versa~\cite{dlb-ictai96,kas-iccad02,HLO07}. This section focus
solely on weighted partial MaxSAT, since the encodings of the other
variants easily follow.

The encoding of hard clauses from weighted partial MaxSAT to PBO is
straightforward, since propositional clauses are a particular case of
pseudo-Boolean constraints. However, for each soft clause 
$\omega_i = (l_1 \vee l_2 \vee \ldots \vee l_k)$ with weight $c_i$, 
the encoding to PBO involves the use of an additional selection
variable $s_i$, such that the corresponding constraint in PBO to 
$\omega_i$ would be $s_i + \sum_{j=1}^k l_j \ge 1$. This ensures that
variable $s_i$ is assigned to true whenever $\omega_i$ is not satisfied.
The objective function of the corresponding PBO instance is to
minimize the weighted sum of the selection variables. For each
selection variable $s_i$ in the objective function, its coefficient
is the weight $c_i$ of the corresponding soft clause $\omega_i$.

\begin{example}
Consider the following weighted partial MaxSAT instance.
\begin{equation}
  \begin{array}{rrl}
    \varphi_h &= \{ & (x_1 \vee x_2 \vee \bar{x}_3), (\bar{x}_2 \vee x_3),
    (\bar{x}_1 \vee x_3) \} \\
    \varphi_s &= \{ & (\bar{x}_3, 6), (x_1 \vee x_2, 3), (x_1 \vee x_3, 2) \} \\
  \end{array}
  \label{eq:wms_instance}
\end{equation}
According to the described encoding, the corresponding PBO instance would be:
\begin{equation}
  \begin{array}{rrl}
    \mbox{minimize } & 6 s_1 + 3 s_2 + 2 s_3 \\
    \mbox{subject to } & x_1 + x_2 + \bar{x}_3 \ge 1 \\
    & \bar{x}_2 + x_3 \ge 1 \\
    & \bar{x}_1 + x_3 \ge 1 \\
    & s_1 + \bar{x}_3 \ge 1 \\
    & s_2 + x_1 + x_2 \ge 1 \\
    & s_3 + x_1 + x_3 \ge 1 \\
  \end{array}
  \label{eq:pbo_instance}
\end{equation}
\end{example}

The encoding of PBO constraints into MaxSAT can be done using any
of the proposed encodings from pseudo-Boolean constraints to 
clauses~\cite{warners-ipl98,roussel-jsat06,een-jsat06}. Hence, 
for each pseudo-Boolean constraint there 
will be a set of hard clauses encoding it in the respective MaxSAT
instance. The number of clauses and additional variables, depends 
on the translation process used. The encoding is trivial when 
the original constraint in the PBO instance is already a clause.

The objective function of PBO instances can be encoded into MaxSAT
with the use of weighted soft clauses. The idea is that for each 
variable $x_j$ with coefficient $c_j$ in the objective function, a 
corresponding soft clause $(\bar{x}_j)$ with weight $c_j$ is 
added to the MaxSAT instance. Therefore, the solution
of the MaxSAT formulation minimizes the weighted sum of problem
variables, as required in the PBO instance.

\begin{example}
For illustration purposes, consider the following PBO instance:
\begin{equation}
  \begin{array}{rrl}
    \mbox{minimize } & 4 x_1 + 2 x_2 + x_3 \\
    \mbox{subject to } & 2 x_1 + 3 x_2 + 5 x_3 \ge 5 \\
    & \bar{x}_1 + \bar{x}_2 \ge 1 \\
    & x_1 + x_2 + x_3 \ge 2 \\
  \end{array}
  \label{eq:pbo_instance2}
\end{equation}
Note that the first and third constraint must be encoded into CNF,
but the second constraint is already a clause and so it can be
represented directly as a hard clause. The corresponding MaxSAT
instance would be:
\begin{equation}
  \begin{array}{rrl}
    \varphi_h &= \{ & \mbox{CNF}(2 x_1 + 3 x_2 + 5 x_3 \ge 5), 
    (\bar{x}_1 \vee \bar{x}_2), \mbox{CNF}(x_1 + x_2 + x_3 \ge 2) \} \\
    \varphi_s &= \{ & (\bar{x}_1, 4), (\bar{x}_2, 2), (\bar{x}_3, 1) \} \\
  \end{array}
  \label{eq:wms_instance2}
\end{equation}
\end{example}

\subsection{Unsatisfiability-Based MaxSAT}

Recent work proposed the use of SAT solvers to solve (partial) MaxSAT,
by iteratively identifying and relaxing unsatisfiable
sub-formulas~\cite{FM06,jpms-date08,jpms-corr07,jpms-sat08b}. In this
paper we refer to these algorithms generically as MSU (Maximum
Satisfiability with Unsatisfiability) algorithms. 

The original algorithm of Fu\&Malik (referred to as MSU1.0)
iteratively identifies unsatisfiable sub-formulas. For each computed
unsatisfiable sub-formula, all original (soft) clauses are relaxed
with fresh relaxation variables. Moreover, a new Equals1 (or AtMost1)
constraint relates the relaxation variables of each iteration,
i.e. exactly 1 of these relaxation variables can be assigned value 1.
The MSU1.0 algorithm can use more than one relaxation variables for each
clause. In the original algorithm~\cite{FM06}, a quadratic pairwise
encoding of the Equals1 constraint was used. Finally, observe that the
Equals1 constraint in line~\ref{li:msu:card} of
Algorithm~\ref{alg:msu1} can be replaced by an AtMost1 constraint,
without affecting the correctness of the algorithm.

More recently, several new MSU algorithms were
proposed~\cite{jpms-corr07,jpms-date08}. The differences of the MSU
algorithms include the number of cardinality constraints used, the
encoding of cardinality constraints (of which the AtMost1 and Equals1
constraints are a special case), the number of relaxation variables
considered for each clause, and how the MSU algorithm
proceeds. Extensive experimentation (from~\cite{jpms-sat08b} but also
from the MaxSAT Evaluation~\cite{maxsat-evaluation})
suggests that an optimized variation of Fu\&Malik's
algorithm\cite{jpms-sat08b} is currently the best performing MSU
algorithm.

\begin{algorithm}[t]
{\small
\begin{codebox}
\Procname{$\proc{msu1}(\varphi)$}
\li $\varphi_{W} \gets \varphi$ \RComment Working formula, initially
set to $\varphi$
\li \While \kw{true}\label{li:msu:init-loop}
\li   \Do $(\textnormal{st}, \varphi_C) \gets \textsc{SAT}(\varphi_{W})$\label{li:msu:satcall}
\li   \Comment{$\varphi_C$ is an unsatisfiable sub-formula if $\varphi_W$ is unsat}
\li   \If $\textnormal{st} = \kw{UNSAT}$
\li     \Then\label{li:msu:unsat}
           $V_R \gets \emptyset$
\li        \For\kw{each} $\omega\in\varphi_C$
\li           \Do
              \If \kw{not} hard($\omega$)
\li              \Then 
                    $r$ is a new relaxation variable
\li                 $\omega_R \gets \omega \cup \{ r \}$ \label{li:msu:newbv} \RComment $\omega_R$ is tagged non-auxiliary
\li                 $\varphi_W \gets\varphi_W - \{ \omega \}\cup \{ \omega_R \}$
\li                 $V_R \gets V_R \cup \{ r \}$ \End\End
\li        $\varphi_R \gets \textnormal{CNF}(\sum_{r\in V_R} r = 1)$\label{li:msu:card}
\RComment Equals1 constraint
\li        Set all clauses in $\varphi_R$ as hard clauses
\li        $\varphi_W \gets \varphi_W \cup \varphi_R$ \RComment Clauses in $\varphi_R$ are declared hard
\li     \Else\label{li:msu:sat} \Comment{Solution to MaxSAT problem}
\li        $\nu\gets |\,\textnormal{blocking variables w/ value 1}\,|$
\li        \Return $|\varphi| - \nu$ \End\label{li:msu:sol}
\End
\end{codebox}
}
\caption{The (Partial) MaxSAT algorithm of Fu\&Malik~\cite{FM06}}
\label{alg:msu1}
\end{algorithm}

\section{Unsatisfiability-Based Weighted MaxSAT}
\label{sec:wmxsat}

This section describes extensions of MSU1.X, described in
Algorithm~\ref{alg:msu1}, for solving (Partial) Weighted MaxSAT
problems.
One simple solution is to create $c_j$ replicas of clause $\omega_j$,
where $c_j$ is the weight of clause $\omega_j$. The resulting extended
CNF formula can then be solved by MSU1.X.
The proof of Fu\&Malik's paper would also apply in this case, and so
correctness follows.
The operation of this solution for (Partial) Weighted MaxSAT justifies
a few observations. Consider an unsatisfiable sub-formula $\varphi_C$
where the smallest weight is ${\mathit{min}}_c$. Each clause would be
replaced by a number of replicas. Hence, this unsatisfiable
sub-formula would be identified ${\mathit{min}}_c$ times.
Clearly, this solution is unlikely to scale for clauses with very
large weights. Hence, a more effective solution is needed, which is
detailed below.

\begin{algorithm}[t]
{\small
\begin{codebox}
\Procname{$\proc{wmsu1}(\varphi)$}
\li $\varphi_{W} \gets \varphi$ \RComment Working formula, initially
set to $\varphi$
\li $\mathit{cost}_{\mathit{lb}} \gets 0$
\li \While \kw{true}\label{li:wmsu:init-loop}
\li   \Do $(\textnormal{st}, \varphi_C) \gets \textsc{SAT}(\varphi_{W})$\label{li:wmsu:satcall}
\li   \Comment{$\varphi_C$ is an unsatisfiable sub-formula if $\varphi_W$ is unsat}
\li   \If $\textnormal{st} = \kw{UNSAT}$
\li     \Then\label{li:wmsu:unsat}
          $\mathit{min}_c\gets \infty$
\li        \For\kw{each} $\omega\in\varphi_C$
\li           \Do
              \If \kw{not} hard($\omega$) \kw{and}
              $\mathit{cost}(\omega) < \mathit{min}_c$
\li              \Then
                    $\mathit{min}_c \gets \mathit{cost}(\omega)$ \End\End
\li           $\mathit{cost}_{\mathit{lb}} \gets \mathit{cost}_{\mathit{lb}} + \mathit{min}_c$
\li           $V_R \gets \emptyset$
\li        \For\kw{each} $\omega\in\varphi_C$
\li           \Do
              \If \kw{not} hard($\omega$)
\li              \Then 
                    $r$ is a new relaxation variable
\li                 $V_R \gets V_R \cup \{ r \}$
\li                 $\omega_R \gets \omega \cup \{ r \}$ \label{li:wmsu:newbv} \RComment $\omega_R$ is tagged non-auxiliary
\li                 $\mathit{cost}(\omega_R)\gets \mathit{min}_c$
\li                 \If $\mathit{cost}(\omega) > \mathit{min}_c$
\li                    \Then
                          $\varphi_W \gets\varphi_W \cup \{ \omega_R \}$
\li                       $\mathit{cost}(\omega)\gets \mathit{cost}(\omega)-\mathit{min}_c$
\li                    \Else
                          $\varphi_W \gets\varphi_W - \{ \omega \}\cup \{ \omega_R \}$\End \End\End
\li        $\varphi_R \gets \textnormal{CNF}(\sum_{r\in V_R} r = 1)$\label{li:wmsu:card}
\RComment Equals1 constraint
\li        Set all clauses in $\varphi_R$ as hard clauses
\li        $\varphi_W \gets \varphi_W \cup \varphi_R$ \RComment Clauses in $\varphi_R$ are declared hard
\li     \Else\label{li:wmsu:sat} \Comment{Solution to Weighted MaxSAT problem}
\li        \Return $\mathit{cost}_{\mathit{lb}}$ \End\label{li:wmsu:sol}
\End
\end{codebox}
}
\caption{Unsatisfiability-based (Partial) Weighted MaxSAT algorithm}
\label{alg:wmsu1}
\end{algorithm}

An alternative solution is to split a clause only when the clause is
included in an unsatisfiable sub-formula. The way the clause is split
depends on its weight.
An algorithm implementing this solution is shown in
Algorithm~\ref{alg:wmsu1}. For each unsatisfiable sub-formula, the
smallest weight $\mathit{min}_c$ of the clauses in the sub-formula is
computed. This smallest weight is then used to update a lower bound on
minimum cost of unsatisfiable clauses. Clauses in the unsatisfiable
sub-formula are relaxed. However, if the weight of a clause is larger
than $\mathit{min}_c$, then the clause is split: a new relaxed clause
with weight $\mathit{min}_c$ is created, and the weight of the original
clause is decreased by $\mathit{min}_c$.

\begin{example}
Consider the partial MaxSAT instance in~(\ref{eq:wms_instance}).
Assume that the unsatisfiable sub-formula detected in line~4 of
Algorithm~\ref{alg:wmsu1} is:
\begin{eqnarray}
  \varphi_C & = & \{ \, (\bar x_2 \vee x_3), (\bar x_1 \vee x_3), \quad 
  (\bar x_3, 6), (x_1 \vee x_2, 3) \, \}.
\end{eqnarray}
Then, the smallest weight $min_c$ is 3, and the new formula becomes
$\varphi_W = \varphi_h \cup \varphi_s$, where 
\begin{equation}
  \begin{array}{rll}
    \varphi_h &= \{ & (x_1 \vee x_2 \vee \bar{x}_3), (\bar{x}_2 \vee x_3),
    (\bar{x}_1 \vee x_3), \mbox{CNF}(s_1 + s_2 = 1) \, \} \\
    \varphi_s &= \{ & (\bar{x}_3, 3), (x_1 \vee x_3, 2),
    (s_1 \vee \bar{x}_3, 3), (s_2 \vee x_1 \vee x_2, 3) \, \}.    
  \end{array}
\end{equation}
\end{example}

Observe that the new algorithm can be viewed as a direct optimization
of the naive algorithm outlined earlier. The main difference is that
each iteration of the algorithm collapses $\mathit{min}_c$ iterations
of the naive algorithm. For clauses with large weights the difference
can be significant.

\begin{theorem}[Correctness of WMSU1]
The value returned by Algorithm~\ref{alg:wmsu1} is minimum cost of
non-satisfied clauses in $\varphi$.
\end{theorem}

\begin{proof}
The previous discussion and the proof in~\cite{FM06}.
\end{proof}

\section{Weighted Boolean Optimization}
\label{sec:wbo}

This section introduces Weighted Boolean Optimization (WBO), a new
framework for modeling with hard and soft pseudo-Boolean
constraints, that extends both MaxSAT and its variants and PBO.
Furthermore, a new algorithm based on identifying unsatisfiable
sub-formulas is also proposed for solving WBO.

An Weighted Boolean Optimization (WBO) formula $\varphi$ is composed
of two sets of pseudo-Boolean constraints, $\varphi_s$ and $\varphi_h$,
where $\varphi_s$ contains the soft constraints and  
$\varphi_h$ contains the hard constraints. For each soft constraint
$\omega_i \in \varphi_s$ there is an associated integer weight 
$c_i > 0$. 
The WBO problem consists in finding an assignment to the problem
variables such that all hard constraints are satisfied and the
total weight of the unsatisfied soft constraints is minimized
(i.e. the total weight of satisfied soft constraints is maximized).

It should be noted that WBO represents a generalization of weighted
partial MaxSAT by introducing the use of pseudo-Boolean constraints
instead of just using propositional clauses. Hence, more compact
formulations can be obtained with WBO than with MaxSAT.
Moreover, PBO formulations can also be linearly encoded into WBO.
Constraints in PBO can be directly encoded as hard constraints
in WBO and the objective function can also be encoded as described 
in section~\ref{sec:ms-pbo}.
Therefore, WBO is a generalization of MaxSAT and its variants, as well
as of PBO, allowing a unified modeling framework to integrate both of
these Boolean optimization problems.

\subsection{Unsatisfiability-Based WBO}
\label{sec:wbo_alg}

This section describes how Algorithm~\ref{alg:wmsu1} 
(introduced in Section~\ref{sec:wmxsat}) for weighted partial MaxSAT
can be modified for solving WBO formulas. 
First of all, in a WBO formula, constraints are not restricted
to be propositional clauses. Both soft and hard constraints can
be pseudo-Boolean constraints. Hence, $\varphi$ is a pseudo-Boolean
formula, instead of a CNF formula. Moreover, the use of a SAT solver
in line 6 is replaced with a pseudo-Boolean solver extended with 
the ability to generate an unsatisfiable sub-formula from the original
pseudo-Boolean formula.

Next, if the formula is unsatisfiable, the weight associated
with the unsatisfiable sub-formula is computed in the same way
(lines 9-13) and the soft constraints in the core must also be relaxed
using new relaxation variables (lines 15-24). Consider that
$\omega = \sum a_j l_j \ge b$ denotes the pseudo-Boolean constraint 
to be relaxed using variable $r$. The resulting relaxed constraint 
in line 19 will be $\omega_R = b \cdot r + \sum a_j l_j \ge b$.

Finally, the constraint on the new relaxation variables in line 25 
does not need to be encoded into CNF. The pseudo-Boolean constraint
$\sum_{r \in V_R} r = 1$ can be directly added to $\varphi_W$,
resulting in a more compact formulation, in particular if the
number of soft constraints in the core is large.

In some cases, for an unsatisfiable sub-formula with $k$ soft
constraints, it is possible to use less than $k$ additional
variables. Consider the following soft constraints 
$\omega_1 = \sum_{l_j \in L_1} a_{1j} l_j \ge b_1$ and 
$\omega_2 = \sum_{l_j \in L_2} a_{2j} l_j \ge b_2$ in a given 
unsatisfiable sub-formula, where $L_1$ and $L_2$ denote respectively 
the set of literals in constraints $\omega_1$ and $\omega_2$. 
Additionally, let $x_k \in L_1$, $\bar{x}_k \in L_2$, $a_{1k} \ge b_1$ 
and $a_{2k} \ge b_2$, i.e. assigning $x_k$ to true satisfies $\omega_1$ 
and assigning $x_k$ to false satisfies $\omega_2$.\footnote{This is a 
  generalization to pseudo-Boolean constraints.
  Note that if the WBO instance corresponds to a MaxSAT instance, 
  this is very common to occur, since $\omega_1$ and $\omega_2$ are
  clauses.}
In this case, these constraints can share the same relaxing variable.
This is due to the fact that it is impossible for both $\omega_1$ and
$\omega_2$ to be unsatisfied by the same assignment, since either 
$x_k$ satisfies $\omega_1$ or $\bar{x}_k$ satisfies $\omega_2$.
Therefore, by using the same relaxing variable on both constraints,
it is maintained the restriction that at most one soft constraint 
in the core can be relaxed.

\begin{example}
Suppose that the following set of soft constraints defines an
unsatisfiable sub-formula in a WBO instance:
\begin{equation}
  \begin{array}{rrl}
    \omega_1 = & 2 x_1 + 3 x_2 + 5 x_3 & \ge 5 \\
    \omega_2 = & \bar{x}_1 + \bar{x}_2 & \ge 1 \\
    \omega_3 = & x_2 + \bar{x}_3 & \ge 1 \\
    \omega_4 = & x_1 + \bar{x}_3 & \ge 1 \\
  \end{array}
  \label{eq:wbo_unsat_core}
\end{equation}
In this case, constraints $\omega_1$ and $\omega_3$ can share the
same relaxation variable, since the assignment of a value to $x_3$ 
implies that either $\omega_1$ or $\omega_3$ is satisfied. 
The same occurs with $\omega_2$ and $\omega_4$, given that the
assignment to $x_1$ either satisfies $\omega_2$ or $\omega_4$. 
Therefore, after the relaxation, the resulting formula can include
just two relaxation variables, instead of four. The resulting 
formula would be:
\begin{equation}
  \begin{array}{rl}
    5 s_1 + 2 x_1 + 3 x_2 + 5 x_3 & \ge 5 \\
    s_2 + \bar{x}_1 + \bar{x}_2 & \ge 1 \\
    s_1 + x_2 + \bar{x}_3 & \ge 1 \\
    s_2 + x_1 + \bar{x}_3 & \ge 1 \\
    s_1 + s_2 & \le 1\\
  \end{array}
  \label{eq:wbo_unsat_core_relax}
\end{equation}
\end{example}

The application of this reduction rule of relaxing variables raises
the problem of finding the smallest number of relaxation variables to
be used. This problem can be mapped into finding a matching of
maximum cardinality in an undirected graph. In such a graph,
there is a vertex for each constraint in the unsatisfiable
sub-formula, while edges connect vertexes corresponding to constraints
that can share a relaxation variable. The problem of finding a
matching of maximum cardinality in an undirected graph can be solved in
polynomial time~\cite{edmonds-65}.
Nevertheless, our prototype implementation of WBO solver uses a 
greedy algorithmic approach.

\subsection{Other Algorithms for WBO}

An alternative solution for solving WBO is to extend existing PBO
algorithms. For example, soft pseudo-Boolean constraints can be 
represented in a PBO instance as relaxable constraints, and the 
overall cost function becomes the weighted sum of the relaxation 
variables of all soft pseudo-Boolean constraints of the original 
WBO formulation.
This solution resembles the existing approach for solving MaxSAT with
PBO~\cite{dlb-ictai96,kas-iccad02}, and has the same potential
drawbacks.

One additional alternative solution is to generalize branch and bound
weighted partial MaxSAT solvers to deal with soft and hard 
pseudo-Boolean constraints. However, note that these approaches focus 
on a search process that uses successive refinements on the
upper bound of the WBO solution, while the algorithm proposed in
section~\ref{sec:wbo_alg} works by refining lower bounds on the
optimum solution value.

\section{Results}
\label{sec:res}

\begin{comment}
We have performed an experimental investigation over a set of Weighted 
\textsc{MaxSat} solvers. The solvers are: MiniMaxSat~\cite{HLO08}, 
IncWMaxSatz~\cite{lin-ijcai07}, W-MaxSatz~\cite{argelich-ncp07} and 
Clone~\cite{pipatsrisawat08solving}. 
The latter is removed from the results because has memory problems
to load the instances, even incrementing the java memory allocation pool.
Experimentation has been performed in a Linux AMD Opteron 2GHz with
1MB of RAM.
\end{comment}

With the objective of evaluating the new (partial) weighted MaxSAT 
algorithm and the new WBO solver, a set of industrially-motivated 
problem instances was selected. The characteristics of the 
classes of instances considered are shown in Table~\ref{tab:inst}. 
For each class of instances, the table provides the class name, 
the number of instances (\#I), the type of MaxSAT variant, and the 
source for the class of 
instances.

\begin{table}[t]
\begin{center}
\caption{Classes of problem instances}
\label{tab:inst}
\begin{tabular}{|c|r|l|l|} \hline
Class & \#I & MaxSAT Variant   & Source \\ \hline\hline
IND   & 110 & Partial Weighted & To Appear in MaxSAT Evaluation 2009 \\ \hline
FIR   & 59  & Partial          & Pseudo-Boolean Evaluation 2007 \\ \hline
SYN   & 74  & Partial          & Pseudo-Boolean Evaluation 2005 \\ \hline
\end{tabular}
\end{center}
\end{table}

Moreover, a wide range of MaxSAT and PBO solvers were considered, all
among the best performing in either the MaxSAT or the Pseudo-Boolean 
evaluations.
The weighted MaxSAT solvers considered were
WMaxSatz~\cite{argelich-ncp07}, MiniMaxSat~\cite{HLO08},
IncWMaxSatz~\cite{lin-ijcai07}, Clone~\cite{pipatsrisawat08solving},
and SAT4J (MaxSAT)~\cite{sat4j}.
In addition, a new version of
MSUnCore~\cite{jpms-corr07,jpms-date08,jpms-sat08b}, integrating the
weighted MaxSAT algorithm proposed in Section~\ref{sec:wmxsat}, was
also evaluated.
The PBO solvers considered were BSOLO~\cite{jpms-date05},
PBS~\cite{kas-iccad02}, Pueblo~\cite{kas-jsat06}, 
Minisat+\cite{een-jsat06}, and SAT4J (PB)~\cite{sat4j}.
Finally, results for the new WBO solver, implementing the WBO 
organization described in Section~\ref{sec:wbo} is also shown.

All experiments were run on a cluster of Linux AMD Opteron 2GHz
servers with 1GB of RAM. The CPU time limit was set to 1800 seconds,
and the RAM limit was set to 1 GB.

All algorithms were run on all problem instances considered. The
original representations were used, in order to avoid introducing any
bias towards any of the problem representations. Tables~\ref{tab:abmx}
and~\ref{tab:abpb} summarize the number of instances aborted by each
solver for each class of instances. As can be concluded, for
practical problem instances, only a small number of MaxSAT solvers is
effective. The results are somewhat different for the PBO solvers,
where several can be competitive for different classes of instances.
It should be noted that the IND benchmarks can be considered
challenging for pseudo-Boolean solvers due to the large clause 
weights used.

For class IND and for the MaxSAT solvers, the results are somewhat
surprising. Some of the solvers perform extremely well, whereas the
others cannot solve most of the problem instances. IncWMaxSatz,
MSUnCore and WBO are capable of solving {\em all} problem instances,
but other MaxSAT solvers abort the vast majority of the problem
instances.
One additional observation is the very good performance of IncWMaxSatz
when compared to WMaxSatz. This clearly indicates that the 
lower bound computation
used in IncWMaxSatz can be very effective, even for industrial problem
instances.
For the PBO solvers, given the set of benchmark instances considered,
SAT4J (PB) and BSOLO come out as the best performing. Clearly, this
conclusion is based on the class of instances considered, which
nevertheless derive from practical applications.
Moreover, SAT4J (PB) performs significantly better than SAT4J
(MaxSAT). This may be the result of a less effective encoding 
internally to SAT4J.

\begin{table}[t]
\begin{center}
\caption{Solved Instances for MaxSAT Solvers}
\label{tab:abmx}
\begin{tabular}{|l|r|r|r|r|r|r|} \hline
Class & WMaxSatz & MiniMaxSat & IncWMaxSatz & Clone & SAT4J (MS) & MSUncore \\ \hline \hline
IND   &    11    &      0      &     110    &    0  &   10       & 110  \\ \hline
FIR   &     7    &     14      &      33    &    5  &   10       &  45  \\ \hline
SYN   &    22    &     29      &      19    &   13  &   21       &  34  \\ \hline \hline
Total
(Out of
243)  &   40     &     43      &     162    &   18  &   41       & 189  \\ \hline
\end{tabular}
\end{center}
\end{table}

\begin{table}[t]
\begin{center}
\caption{Solved Instances for PBO \& WBO Solvers}
\label{tab:abpb}
\begin{tabular}{|l|r|r|r|r|r|r|} \hline
Class & BSOLO &  PBS  & Pueblo & Minisat+ & SAT4J (PB) & WBO  \\ \hline \hline
IND   &  17   &    0  &    0   &     0    &  60        & 110  \\ \hline
FIR   &  20   &   11  &   14   &    22    &   7        &  39  \\ \hline
SYN   &  51   &   19  &   30   &    30    &  22        &  33  \\ \hline \hline
Total
(Out of
243)  &  88   &   30  &   44   &    52    &   89       & 172  \\ \hline
\end{tabular}
\end{center}
\end{table}

\begin{comment}
\begin{table}[t]
\begin{center}
\caption{Aborted Instances for MaxSAT \& WBO Solvers}
\label{tab:abmx}
\begin{tabular}{|l|r|r|r|r|r|r|} \hline
Class & WMaxSatz & MiniMaxSat & IncWMaxSatz & Clone & SAT4J & MSUncore \\ \hline \hline
IND   &   99     &    110      &      0      &  110  &  100  &  0  \\ \hline
FIR   &   52     &     45      &      33     &   54  &   49  & 14  \\ \hline
SYN   &   52     &     45      &      55     &   61  &   53  & 40  \\ \hline \hline
Total (Out
of 243&  203     &    200      &      88     &  225  &  202  & 54  \\ \hline
\end{tabular}
\end{center}
\end{table}

\begin{table}[t]
\begin{center}
\caption{Aborted Instances for PBO \& WBO Solvers}
\label{tab:abpb}
\begin{tabular}{|l|r|r|r|r|r|r|} \hline
Class & BSOLO &  PBS  & Pueblo & Minisat+ & SAT4J & WBO \\ \hline \hline
IND   &  93   &       &  110   &   110    &  50   &  0  \\ \hline
FIR   &  39   &       &   45   &    37    &  52   & 20  \\ \hline
SYN   &  23   &       &   44   &    44    &  52   & 41  \\ \hline \hline
Total &  155  &       &  199   &   191    &  154  & 61  \\ \hline
\end{tabular}
\end{center}
\end{table}
\end{comment}

\begin{figure}[t]
  \scalebox{1.1}{\begin{picture}(0,0)\includegraphics{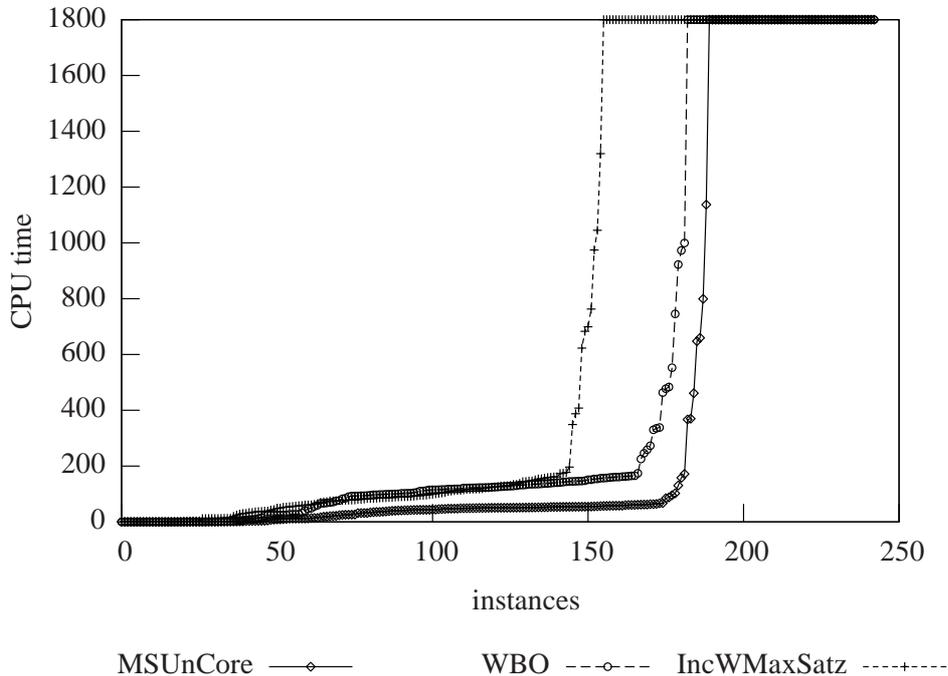}\end{picture}\begingroup
\setlength{\unitlength}{0.0200bp}\begin{picture}(16200,11880)(0,0)\put(1925,2750){\makebox(0,0)[r]{\strut{} 0}}\put(1925,3703){\makebox(0,0)[r]{\strut{} 200}}\put(1925,4657){\makebox(0,0)[r]{\strut{} 400}}\put(1925,5610){\makebox(0,0)[r]{\strut{} 600}}\put(1925,6563){\makebox(0,0)[r]{\strut{} 800}}\put(1925,7517){\makebox(0,0)[r]{\strut{} 1000}}\put(1925,8470){\makebox(0,0)[r]{\strut{} 1200}}\put(1925,9423){\makebox(0,0)[r]{\strut{} 1400}}\put(1925,10377){\makebox(0,0)[r]{\strut{} 1600}}\put(1925,11330){\makebox(0,0)[r]{\strut{} 1800}}\put(2200,2200){\makebox(0,0){\strut{} 0}}\put(4835,2200){\makebox(0,0){\strut{} 50}}\put(7470,2200){\makebox(0,0){\strut{} 100}}\put(10105,2200){\makebox(0,0){\strut{} 150}}\put(12740,2200){\makebox(0,0){\strut{} 200}}\put(15375,2200){\makebox(0,0){\strut{} 250}}\put(550,7040){\rotatebox{90}{\makebox(0,0){\strut{}CPU time}}}\put(9062,1375){\makebox(0,0){\strut{}instances}}\put(4430,275){\makebox(0,0)[r]{\strut{}MSUnCore}}\put(9463,275){\makebox(0,0)[r]{\strut{}WBO}}\put(14496,275){\makebox(0,0)[r]{\strut{}IncWMaxSatz}}\end{picture}\endgroup

}
  \label{fig:incr-time}
  \caption{Run times for IncWMaxSatz, MSUnCore, and WBO for all instances}
\end{figure}

Motivated by the overall results, the best MaxSAT, PBO and the WBO
solver were analyzed in more detail. Given the experimental results,
IncWMaxSatz, MSUnCore, and WBO were
selected. Figure~\ref{fig:incr-time} shows the results for the
selected solvers by increasing run times.

As can be concluded, the plot confirms the trends in the tables of
results. MSUnCore is the best performing, followed by WBO and
IncWMaxSatz. For smaller run times (instances from class IND),
IncWMaxSatz can be more efficient than WBO. Moreover, these results
indicate that, for the classes of instances considered, encoding
cardinality constraints into CNF (as done in MSUnCore) may be a better
solution than natively handling cardinality and pseudo-Boolean
constraints (as done in WBO). 
It should be noted that all the instances considered can be encoded
with cardinality constraints, for which existing polynomial encodings
guarantee arc-consistency. This is not true for problem instances that
use other pseudo-Boolean constraints, and for which encodings that
ensure arc-consistency are exponential in the
worst-case~\cite{een-jsat06}.
Finally, another source of difference in the experimental results is
that whereas MSUnCore is built on top of PicoSAT~\cite{biere-jsat08},
WBO is built on top of Minisat2. The different underlying SAT solvers
may also contribute to explain some of the differences observed.

\section{Related Work} \vspace*{-0.05cm}
\label{sec:relw}

A brief account of MaxSAT and PBO solvers is provided in
Section~\ref{sec:prelim}.
The use of unsatisfiability for solving MaxSAT was first proposed in
2006~\cite{FM06}. This work was later
extended~\cite{jpms-corr07,jpms-date08,jpms-sat08b}, to accommodate
several alternative algorithms and a number of optimizations to the
first algorithm.
To the best of our knowledge, MSUnCore is the first algorithm for
solving (Partial) Weighted MaxSAT with unsatisfiable sub-formula
identification. Also, to the best of our knowledge, WBO represents a
new modeling framework, and the associate algorithm is new.

The use of optimization variants of decision procedures has also been
proposed in the area of SMT~\cite{nieuwenhuis-sat06}, and a few SMT
solvers now offer the ability for solving optimization problems. The
approaches used for solving optimization problems in SMT are based on
the use of relaxation variables, similarly to the PBO approach for
solving MaxSAT~\cite{kas-iccad02}.

\begin{comment}
Observing the results of the last MaxSAT Evaluation~\cite{maxsat-evaluation},
the most remarkable algorithms solving weighted MaxSAT are: 
Clone~\cite{pipatsrisawat08solving}, which applies a branch and bound 
algorithm to a d-DDNF formula, taking the weighting information from
the original MaxSAT instanc;
IncWMaxSatz~\cite{lin-ijcai07}, an extension of unweighted MaxSAT solver 
MaxSatz~\cite{LMP07} to weighted MaxSAT, with the incorporation of 
a new lower bound;
MiniMaxSAT~\cite{HLO08}, an extension of MiniSat+~\cite{een-jsat06} with 
the incorporation of inference rules focussed on dealing with weighted 
and hard clauses;
and SAT4J MAXSAT, an extension of SAT solver SAT4J, which iteratively 
translates the problem to SAT.
\end{comment}

\section{Conclusions and Future Work} \label{sec:conc}

This paper proposes a new algorithm for (Partial) Weighted MaxSAT,
based on unsatisfiable sub-formula identification. In addition, the
paper introduces Weighted Boolean Optimization (WBO), that aggregates
and generalizes PBO and MaxSAT. The paper then shows how
unsatisfiability-based algorithms for (Partial) Weighted MaxSAT can be
extended to WBO. Finally, the paper illustrates how to extend other
algorithms for PBO and MaxSAT to solve WBO.

Experimental results, obtained on a representative set of benchmark
instances shows that the new algorithm for weighted MaxSAT can
outperform other existing algorithms by orders of magnitude. The
experimental results also provide a preliminary (albeit possibly
biased) study on the performance differences between handling
pseudo-Boolean constraints natively and encoding to CNF. Finally, the
paper shows that a general algorithm for WBO can be as efficient as
other dedicated algorithms.

The integration of MaxSAT and PBO into a unique optimization extension
of SAT increases the range of problems that can be solved. It also
allows developing other general purpose algorithms, integrating the
best techniques from both domains.
Future research work will address adapting other algorithms for
WBO. One concrete example is the use of PBO solvers. The other is
extending the existing family of MSU algorithms for WBO.

\vspace{1em}
\noindent
\textbf{Acknowledgement.}
This work is partially supported by EU grant ICT/217069 and 
FCT grant PTDC/EIA/76572/2006.

\bibliography{bibtex-db,extra,maxsat}
\bibliographystyle{abbrv}

\end{document}